\documentclass[conference, letterpaper]{IEEEtran}
\IEEEoverridecommandlockouts

\usepackage{cite}
\usepackage{amsmath,amssymb,amsfonts}
\usepackage{graphicx}
\usepackage{textcomp}
\usepackage{xcolor}

\usepackage[hyphens]{url}
\usepackage{booktabs}
\usepackage{multirow}
\usepackage{dblfloatfix}
\usepackage[noend]{algpseudocode}
\usepackage{algorithm}
\usepackage{here}
\usepackage{bm}
\algnewcommand\algorithmicforeach{\textbf{for each}}
\algdef{S}[FOR]{ForEach}[1]{\algorithmicforeach\ #1\ \algorithmicdo}

\def\BibTeX{{\rm B\kern-.05em{\sc i\kern-.025em b}\kern-.08em
    T\kern-.1667em\lower.7ex\hbox{E}\kern-.125emX}}
\begin{document}

\title{Feasibility Study of Magnetism-based \\ Indoor Positioning Methods in an Incineration Plant}

\author{\IEEEauthorblockN{Rei Okumura, Ismail Arai}
\IEEEauthorblockA{\textit{Nara Institute of Science and Technology} \\
Nara, Japan \\
okumura.rei.os5@is.naist.jp, \\ ismail@itc.naist.jp}
\and
\IEEEauthorblockN{Atarashi Yutaro, Kawabata Kaoru}
\IEEEauthorblockA{Hitachi Zosen Corporation \\
Osaka, Japan \\
atarashi.y@hitachizosen.co.jp, \\ kawabata\_k@hitachizosen.co.jp}
\and
\IEEEauthorblockN{Kazutoshi Fujikawa}
\IEEEauthorblockA{\textit{Nara Institute of Science and Technology} \\
Nara, Japan \\
fujikawa@itc.naist.jp}
}

\maketitle

\pagestyle{empty}


\begin{abstract}
In an incineration plant, remote operation from a centralized control room is now possible, but inspection and cleaning of equipment still require a worker to visit the site.
When the plant owner reduces the number of workers due to operation costs, it will be standard for a single worker to visit the site. Therefore, it is necessary to monitor the location of workers in real-time to detect unexpected human accidents quickly.
Conventional methods use radio waves, such as Wi-Fi and Bluetooth, but there is little demand for communication equipment in the incineration plant.
However, there is not enough demand for communication facilities in the incineration plant.
It is too large to bear the cost of installing wireless access points, and Bluetooth Low Energy (BLE) beacons just for positioning.
Therefore, we are focusing on magnetism using for indoor positioning method.
In addition, the incineration plant has a lot of types of equipment that contains a wide range of magnetized metals, large motors, and generators.
We could observe the magnetic peculiarity at each point.
Based on these assumptions, we have developed a new indoor positioning method at the incineration plant.
This paper describes the development of an indoor positioning system for an incineration plant.
And we propose three methods for fingerprinting matching: Point matching, Path matching, and DTW matching.
The average positioning errors of these methods are 6.89 m, 0.05 m, and 0.06 m, respectively.
\end{abstract}


\section{Introduction}
Production plants and incineration plants have been trying to improve operating efficiency and reduce labor costs by automating their operating systems in recent years. In regular operation, workers can monitor and manage equipment remotely from a centralized control room, enabling safe operation by a small number of people.
However, the equipment and facilities in the plant need to be inspected and cleaned periodically, and it isn't easy to do so remotely. Therefore, workers must visit the site, and it remains a challenge to ensure both worker safety and labor cost reduction. To solve this problem, we are developing a system that can monitor the location of workers in real-time.

\begin{figure}[htbp]
    \centering
\begin{minipage}[b]{.45\linewidth}
    \centering
    \includegraphics[width=\linewidth]{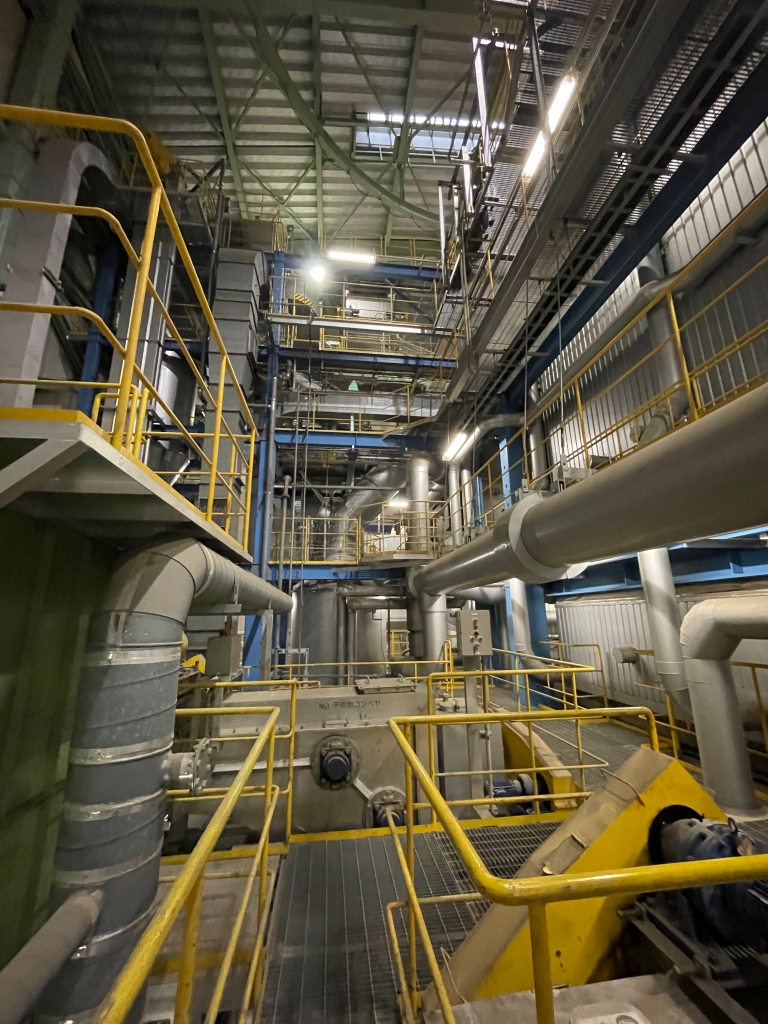}
    \caption{Overview of an incineration.}
    \label{fig:clean_factory}
\end{minipage}
\begin{minipage}[b]{.45\linewidth}
    \centering
    \includegraphics[width=\linewidth]{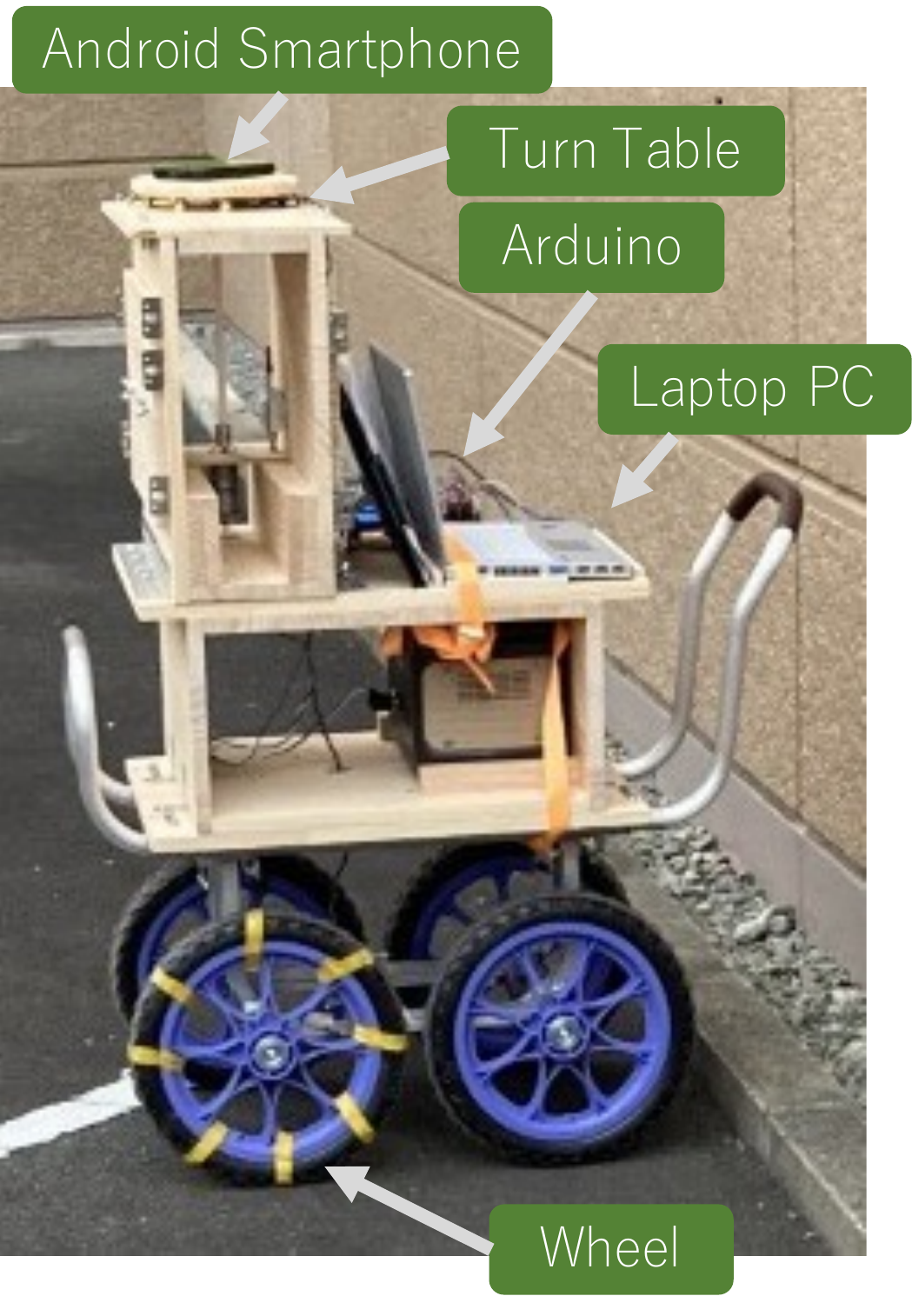}
    \caption{Our measuring cart appearance.}
    \label{fig:cart}
\end{minipage}
\end{figure}

An incineration plant has an ample space called a furnace room. The furnace room has fully automated incineration of incinerated materials arranged in three dimensions.
Figure~\ref{fig:clean_factory} is a photograph of the furnace room. It shows the metal equipment, pipes, and scaffolding constructed with grating.
As it shows, the incineration plant is characterized by its unique furnace room, which occupies a large part of the plant space.
This paper presents the results of a basic experiment for indoor positioning using the measurement cart shown in Figure~\ref{fig:cart} at such an incineration plant.

\section{Literature Review}
The system that monitors the location of workers in an incineration plant in real-time is one of the indoor positioning systems.
Thus, we focus on the indoor positioning system.
Sadowski et al. \cite{Sadowski-rssi} proposed a method using the Received Signal Strength Indicator (RSSI), which is the signal strength of Wi-Fi
access points (APs).
They estimate using the trilateration method for multiple points in two environments, an indoor laboratory and a classroom. The location multiple points are 1 m - 5 m away from the Wi-Fi APs, and an average estimation error is 0.664 m.
On the other hand, they show that the RSSI value changes logarithmically as the distance increases between the Wi-Fi APs and the receiving devices.

Sen et al. \cite{Sen-csi} also proposed using Channel State Information (CSI), which is Multidimensional data of amplitude and phase displacements obtained from routers, collected for each radio wave path.
They found that the estimation error is within 4 m when all five APs are used.

A similar method uses Bluetooth Low Energy (BLE) beacons.
Zhuang et al. \cite{zhuang-2016-bluetooth} have proposed an algorithm for indoor positioning with a smartphone using BLE beacons.
They use four techniques, including channel-separated fingerprinting (FP).
They validate a combination of four techniques, including channel-separated fingerprinting (FP),
and find that the estimation error is less than 2.56 m with a probability of 90~\% on an office environment, which has 20 BLE beacons per 9 m in 60 m $\times$ 40 m research field.
The methods mentioned above require installing Wi-Fi APs and BLE beacons are rarely installed in incineration plants.
It is not realistic to install such infrastructure only for indoor positioning in terms of installation cost considering the operational purpose of the factories.

Magnetic FP is one of the indoor positioning methods that do not require any infrastructure.
Ashraf et al. \cite{Ashraf2020} have proposed a method to classify magnetic data acquired by smartphones using a Convolutional Neural Network (CNN). They use a unique detection algorithm based on magnetic FP to reduce the maximum estimation error to less than 3 m.
FP is a general-purpose algorithm used for indoor positioning, which records the sensor data along with the position (called reference point) in advance to create a data map (Figure~\ref{fig:train_phase}). The center of gravity of the reference points with the closest sensor data or the upper reference points is used as the positioning result (Figure~\ref{fig:estimate_phase}). Using magnetic FP is performed by detecting residual magnetism indoors with a sensor. It has been reported that the positioning accuracy deteriorates when the positioning space becomes large due to the effect that similar magnetism can be observed at multiple reference points \cite{Vandermeulen-2013}.
Therefore, limiting the range of positioning is the key to achieving high accuracy using magnetic FP. In other related papers of compensating for the problems of magnetic positioning,
Higashi et al. \cite{Higashi-2017eg} conducted a study to improve the accuracy by using Wi-Fi.
They have shown the effectiveness of this method. Xu et al.
\cite{Zu-2017} proposed an algorithm to identify the hierarchy of a building by combining barometric pressure and stair ascent/descent detection. They also showed that the algorithm using only barometric pressure could identify the building with 90~\% accuracy.

Our research field is one floor 60 m $\times$ 79 m in size, and there is no precedent for the magnetic FP of this scale to be investigated. Therefore, this paper discusses the problem of magnetic FP under the scenario that the current floor has already been identified using a barometric method with low infrastructure cost.

\section{Magnetic Indoor Positioning Path Matching Method} \label{sec:suggestion}
This study aims to investigate an indoor positioning system that we can implement at the lowest possible cost in an incineration plant.
When we estimated the cost of installing the infrastructure in the study field, we found that it would require funds in the order of one million Japanese yen for BLE beacons and ten million Japanese yen for Wi-Fi APs. Unfortunately, it is not easy for the incineration plant owners to obtain this funding. Furthermore, even if they spent this money to build infrastructure, there are no guarantee existing methods can be used as they are, making it even more unrealistic to build an infrastructure.
We can imagine that the variation of magnetism in the furnace room is a large variety.
Therefore, We expect that the magnetism at any given point has a unique value, and indoor positioning by magnetic FP is possible.

\subsection{Indoor Positioning by Magnetic Sensing}
In indoor positioning using magnetic sensing, the magnetism within the research field is surveyed in advance and matched with the current one during positioning.
We refer to the former survey as the offline phase and the latter as the estimation phase.
After that, information such as magnetism and acceleration is referred to as environmental data.
Each phase and the general procedure are as follows.

\begin{figure}[htbp]
    \centering
    \includegraphics[width=\linewidth]{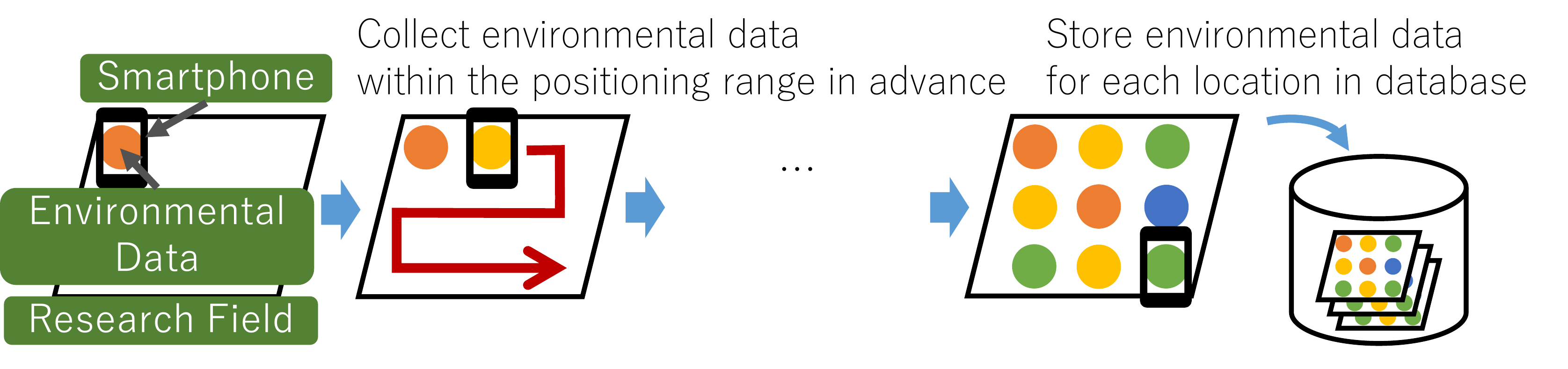}
    \caption{Offline phase.}
    \label{fig:train_phase}
\end{figure}

\begin{figure}[htbp]
    \centering
    \includegraphics[width=\linewidth]{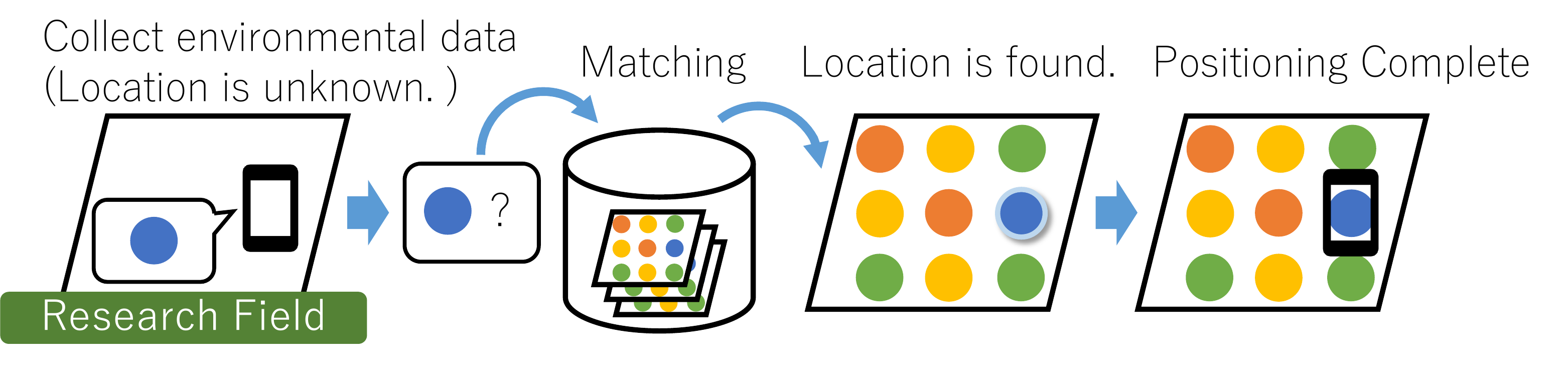}
    \caption{Estimation phase.}
    \label{fig:estimate_phase}
\end{figure}

\begin{itemize}
    \item Offline Phase (Figure~\ref{fig:train_phase})
    \begin{itemize}
        \item Set the reference points in the research field.
        \item Obtain coordinates and environmental data for each reference point.
    \end{itemize}
    \item Estimation Phase (Figure~\ref{fig:estimate_phase})
    \begin{itemize}
        \item Collect environmental data within the research field.
        \item Identify the coordinates by checking them against the reference points obtained in the offline phase.
    \end{itemize}
\end{itemize}

\subsection{Proposed Method in Matching Estimation Phase} \label{ssec:check}
As mentioned above, to improve our system's positioning accuracy, it is important to accurately match the environmental data in the estimation phase to determine the estimated coordinates.

This section describes the specific procedures to be followed in the matching phase.
First, a method for processing magnetic data is presented for matching features, and then three proposed methods for feature matching are described.

\subsubsection{Magnetic Data Processing} \label{ssec:mag_data_preprocessing}
\begin{figure}[htbp]
    \centering
    \includegraphics[width=.9\linewidth]{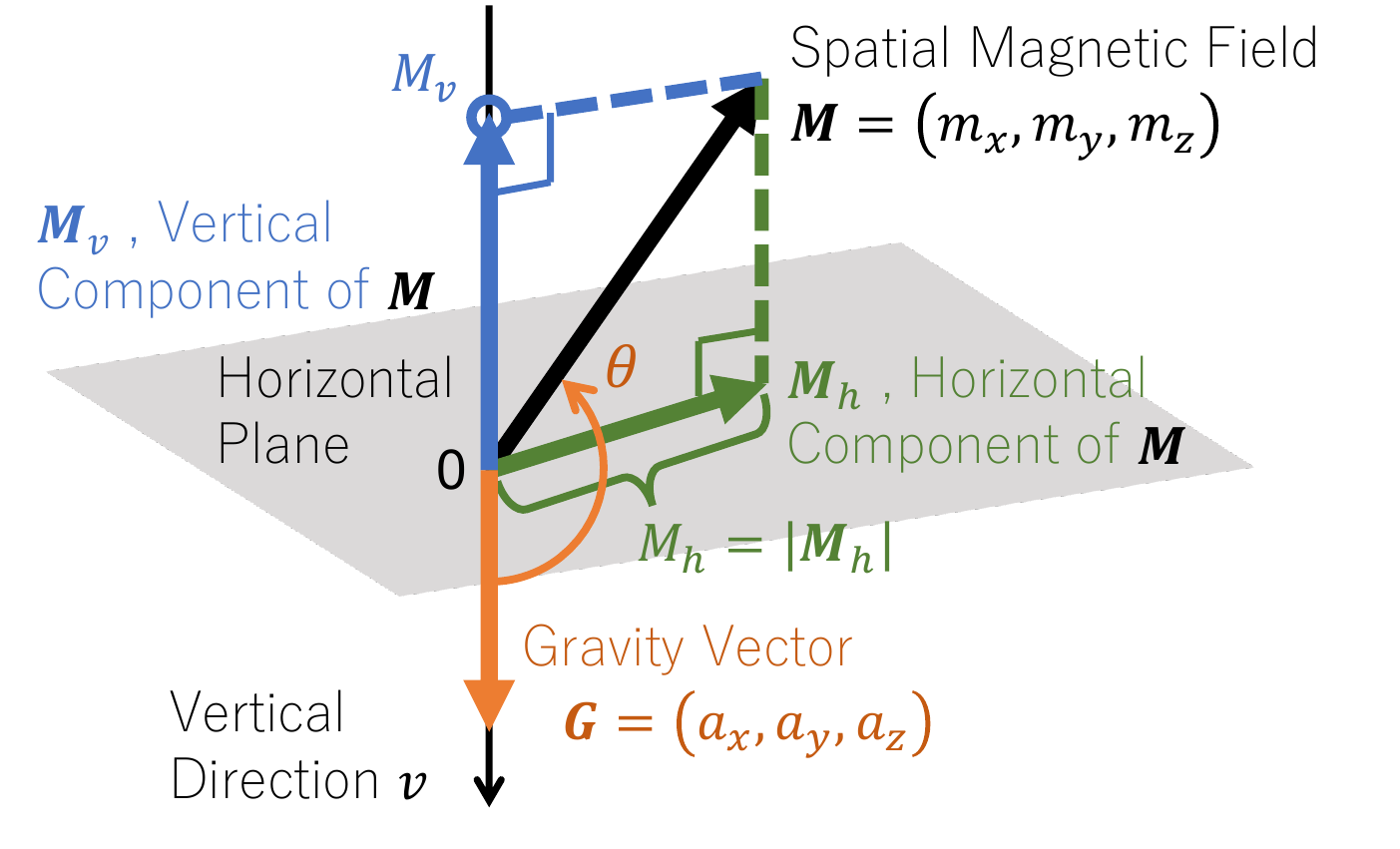}
    \caption{Decomposition of magnetic vectors.}
    \label{fig:mag_hv}
\end{figure}

This section describes $M_v$ and $M_h$, which are feature values during the matching process.
When operating as a positioning system, it is assumed that the orientation of the environmental data collection device differs between the offline phase and the estimation phase. In addition, in the survey field, it is difficult to estimate the device's orientation because the magnetic sensor values required for the orientation estimation are unreliable.
Therefore, it is necessary to use features that take the posture change of the environmental data collection device into account to some extent in the matching process.
Figure~\ref{fig:mag_hv} shows these features, $M_v$ and $M_h$, which are the vertical and horizontal components of the magnetic field applied to the device, respectively. The $M_v$ and $M_h$ are obtained by scalar projection of the magnetic field and gravitational direction used to the device and can be calculated by equations~\eqref{eq:mv} and \eqref{eq:mh} when the output values of the magnetic sensor and accelerometer of the device are $m_x, \ m_y, \ m_z$ and $a_x, \ a_y, \ a_z$.

\begin{align}
    M_v &= \left| \bm{M} \right| \cos\theta \notag \\
        &= \left| \bm{M} \right| \frac{\bm{M} \cdot \bm{G}}{\left| \bm{M} \right| \left| \bm{G} \right|} \notag \\
        &= \frac{\bm{M} \cdot \bm{G}}{\left| \bm{G} \right|}
            = \frac{m_x a_x + m_y a_y + m_z a_z}{\sqrt{a_x^2 + a_y^2 + a_z^2}}
        \label{eq:mv}
\end{align}
\begin{flalign*}
        && \because \bm{M} \cdot \bm{G} = \left| \bm{M} \right| \left| \bm{G} \right| \cos\theta \ \Leftrightarrow \ \cos\theta = \frac{\bm{M} \cdot \bm{G}}{\left| \bm{M} \right| \left| \bm{G} \right|}
\end{flalign*}
\begin{align}
    M_h &= \sqrt{\left| \bm{M} \right|^2 - M_v^2} \label{eq:mh}
\end{align}
\begin{flalign*}
    && \because \left| \bm{M} \right|^2 = M_v^2 + M_h^2
\end{flalign*}

Since the environmental data in this paper was collected with the device screen always facing upward, when the road is horizontal, the real-world world coordinate system with the gravity direction as the Z-axis coincides with the Z-axis of the device coordinate system.
Therefore, the vertical component $M_v$ and horizontal components $M_h$ are obtained
from $m_x, m_y, m_z$ as equations~\eqref{eq:cart_mv} and \eqref{eq:cart_mh}.
The reason for using a measurement cart was to record the position in centimeters accurately.
We employ data obtained from the cart for both phases as a fundamental experiment to avoid extra error factors.
\begin{align}
    M_v &= m_z \label{eq:cart_mv} \\
    M_h &= \sqrt{m_x^2 + m_y^2} \label{eq:cart_mh}
\end{align}

\subsubsection{Point Matching}
This method compares the feature target $M_v$ and $M_h$ as ${\rm tar}(M_v, M_h)$ obtained in the estimation phase with the set of features reference $M_v$ and $M_h$ ${\rm ref}(M_v, M_h)$ obtained in the offline phase. The Euclidean distance is used for comparison, and the reference point with the smallest value is used as the estimation result.

While this algorithm is the simplest and fastest, the number of features is only two, so the estimation error is maybe large.
The estimation algorithm is shown in Algorithm \ref{alg:point_matching}.

\begin{algorithm}[htbp]
    \caption{\ Point Matching}
    \label{alg:point_matching}
    \begin{algorithmic}
        \renewcommand{\algorithmicrequire}{\textbf{Input:}}
        \renewcommand{\algorithmicensure}{\textbf{Output:}}

        \Require \\
        \begin{itemize}
            \item ${\rm ref}(Mv_1, \ Mh_1), \dots, {\rm ref}(Mv_N, \ Mh_N)$

            List of $(M_v, \ M_h)$ collected at offline phase.
            \item ${\rm tar}(Mv, \ Mh)$

            Matching target $(M_v, \ M_h)$.
        \end{itemize}
        \Ensure \\
        \begin{itemize}
            \item $idx$

            Index of the selected reference point.
        \end{itemize}
        \State
    \end{algorithmic}
    \begin{algorithmic}[1]
        \State $minScore \gets 1000000000$ \Comment{Huge number}
        \State $idx \gets 1$
        \State ${\rm t}Mv, {\rm t}Mh \gets {\rm tar}(Mv, \ Mh)$
        \For {$i \gets 1 \ {\rm to} \  N$}
            \State ${\rm r}Mv_i, {\rm r}Mh_i \gets {\rm ref}(Mv_i, \ Mh_i)$
            \State $score \gets \sqrt{({\rm r}Mv_i - {\rm t}Mv)^2 + ({\rm r}Mh_i - {\rm t}Mh)^2}$
            \If {$score < minScore$}
                \State $idx \gets i$
            \EndIf
        \EndFor
        \State \Return $idx$
        \end{algorithmic}
\end{algorithm}

\subsubsection{Path Matching}
\begin{figure}[htbp]
    \centering
    \includegraphics[width=\linewidth]{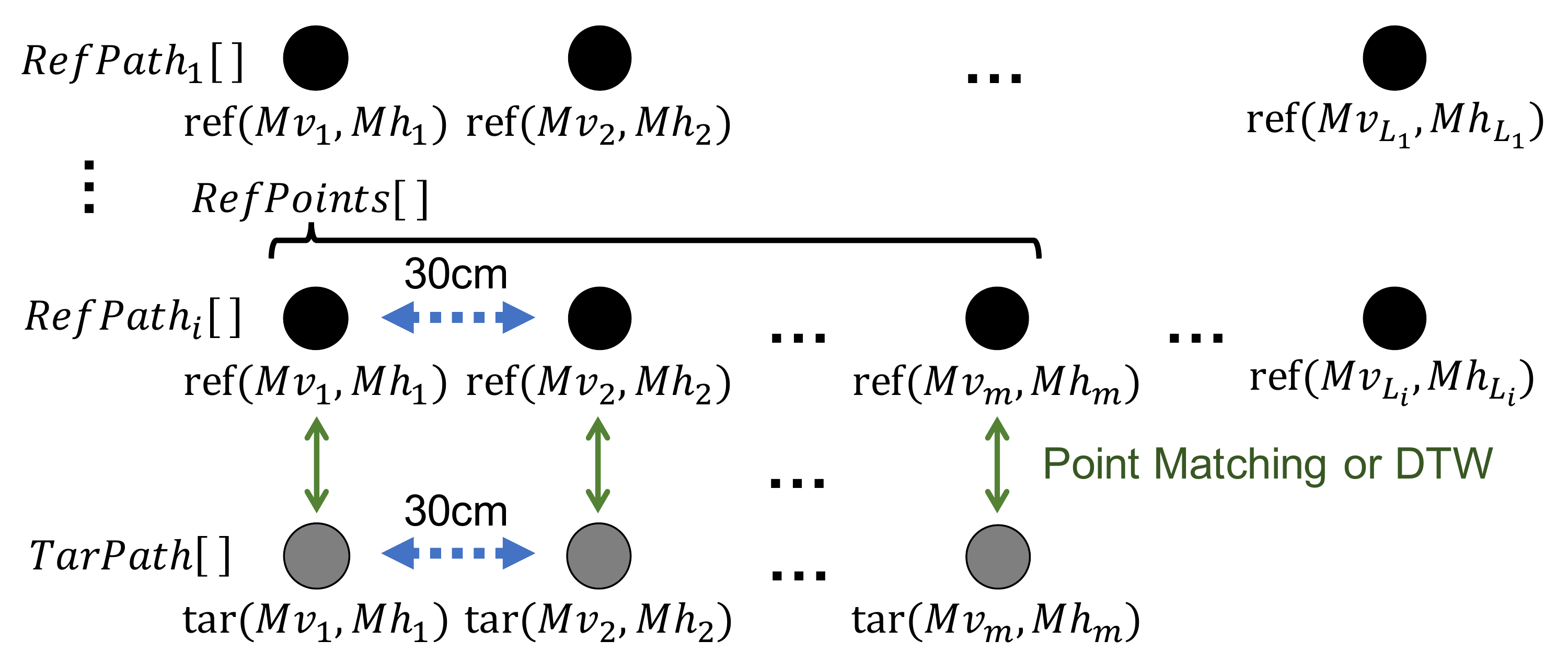}
    \caption{Path Matching.}
    \label{fig:line_matching}
\end{figure}

Considering that a worker walks around the factory, we assume that a path connecting some reference points obtained in the offline phase represents his/her trajectory. Based on this assumption, this method performs multiple points matching along the trajectory. As shown in Figure~\ref{fig:line_matching}, each point of the target path $TarPath[\;]$ obtained in the estimation phase, and the reference path $RefPath[\;]$ obtained in the offline phase is point-matched and compared. The mean value of the scores obtained from each point is used as the score for Path Matching.

This method considers the association with physically close points to each other and is expected to improve the positioning accuracy.
On the other hand, compared to Point Matching, environmental data acquisition requires new information on paths, which are the relationships between reference points, complicates the data collection process, and increases the computation time due to the increased number of matching points.

\begin{algorithm}[htbp]
    \caption{\ Path Matching and DTW Matching}
    \label{alg:path_dtw_matching}
    \begin{algorithmic}
        \renewcommand{\algorithmicrequire}{\textbf{Input:}}
        \renewcommand{\algorithmicensure}{\textbf{Output:}}

        \Require \\
        \begin{itemize}
            \item $RefPaths[\;][\;] = RefPath_1[\;], \dots, RefPath_n[\;]$

            List of path collected at offline phase.
            \item {\small $TarPath[\;] = {\rm tar}(Mv_1, \ Mh_1), \dots, {\rm tar}(Mv_m, \ Mh_m)$}

            List of matching target $(M_v, \ M_h)$.
        \end{itemize}
        \Ensure \\
        \begin{itemize}
            \item $AnsRefPath[\;]$

            Path of estimation result.
            \item $idx$

            How many points in the line should be the starting point?
        \end{itemize}
        \State
    \end{algorithmic}
    \begin{algorithmic}[1]
        \State $minScore \gets 1000000000$ \Comment{Huge number}
        \State $AnsRefPath[\;] \gets RefPaths_1[\;]$
        \State $M \gets {\rm \ length \ of} \ TarPath[\;] $
        \ForAll {$RefPath_j[\;] \in RefPaths[\;][\;]$}
            \State $idx \gets 1$
            \State $L \gets {\rm \ length \ of} \ RefPath_j[\;] $
            \For {$i \gets 1 \ {\rm to} \  L - M$}
                \State $RefPoints[\;] \gets RefPath_j[i \ {\rm to} \ i + M]$
                \If {Path Matching}
                    \State $scores \gets$ Calculate the Euclidean distance between each point in $RefPoints[\;]$ and $TarPath[\;]$.
                    \State $score \gets mean(scores)$ \Comment{Calculate the arithmetic mean.}
                \EndIf
                \If {DTW Matching}
                    \State $score \gets$ Calculate the DTW score in $RefPoints[\;]$ and $TarPath[\;]$.
                \EndIf
                \If {$score < minScore$}
                    \State $AnsRefPath[\;] \gets RefPath[\;]$
                    \State $idx \gets i$
                \EndIf
            \EndFor
        \EndFor
        \State \Return $AnsRefPath[\;]$, $idx$
        \end{algorithmic}
\end{algorithm}

\subsubsection{DTW Matching}
This method employs Dynamic Time Warping (DTW) \cite{senin2008dynamic} instead of the point matching of the Path Matching.
We perform DTW on path $TarPath[\;]$ obtained in the offline phase and path $RefPath[\;]$ obtained in the estimation phase and use the resulting value as the score. Compared with Path Matching, DTW is expected to improve matching accuracy, but it also increases the computational complexity.
Algorithm \ref{alg:path_dtw_matching} shows Path Matching and DTW Matching.

\section{Experimental Setup} \label{sec:experimental_setup}
This section describes how we collected the reference data necessary for the indoor positioning system at the incineration plant.
The environmental data used in this paper was acquired using a cart shown in Figure~\ref{fig:cart}.
In addition, Table~\ref{tab:env_data} shows the list of environmental data items obtained.


\begin{table*}[tb]
    \caption{Smartphone's sensor data}
    \footnotesize
    \begin{tabular}{llll}
        \hline
        Sensor name or collect item & Element count (Axis) or data information & Reason of collected & Comments \\
        \hline \hline
        Geomagnetic field strength [ $\mu$T ] & 3 (XYZ) & Main information of our method & Collected every 30 [ $\mu$s ] \\
        Accelerometer [ m/s${}^2$ ] & 3 (XYZ) & Pose estimation of collection device & Collected every 7 [ $\mu$s ] \\
        Gyroscope [ rad/s ] & 3 (XYZ) & Pose estimation of collection device & Collected every 7 [ $\mu$s ]
    \end{tabular}
    \centering
    \label{tab:env_data}
\end{table*}

A Google Pixel 4a with Android OS version 11 was used.
It is installed at 143.5 cm in the height of the cart.
We installed a self-made Android application with the following functions on the device to collect sensor data.
\begin{enumerate}
    \item The device stores the values which various sensors can collect with timestamps. \label{enu:app:1}
    \item The Device stores the location in the incineration plant with timestamps. \label{enu:app:2}
    \item The measuring cart controls the device via Bluetooth. \label{enu:app:3}
\end{enumerate}

\begin{figure}[tb]
    \centering
    \includegraphics[width=\linewidth]{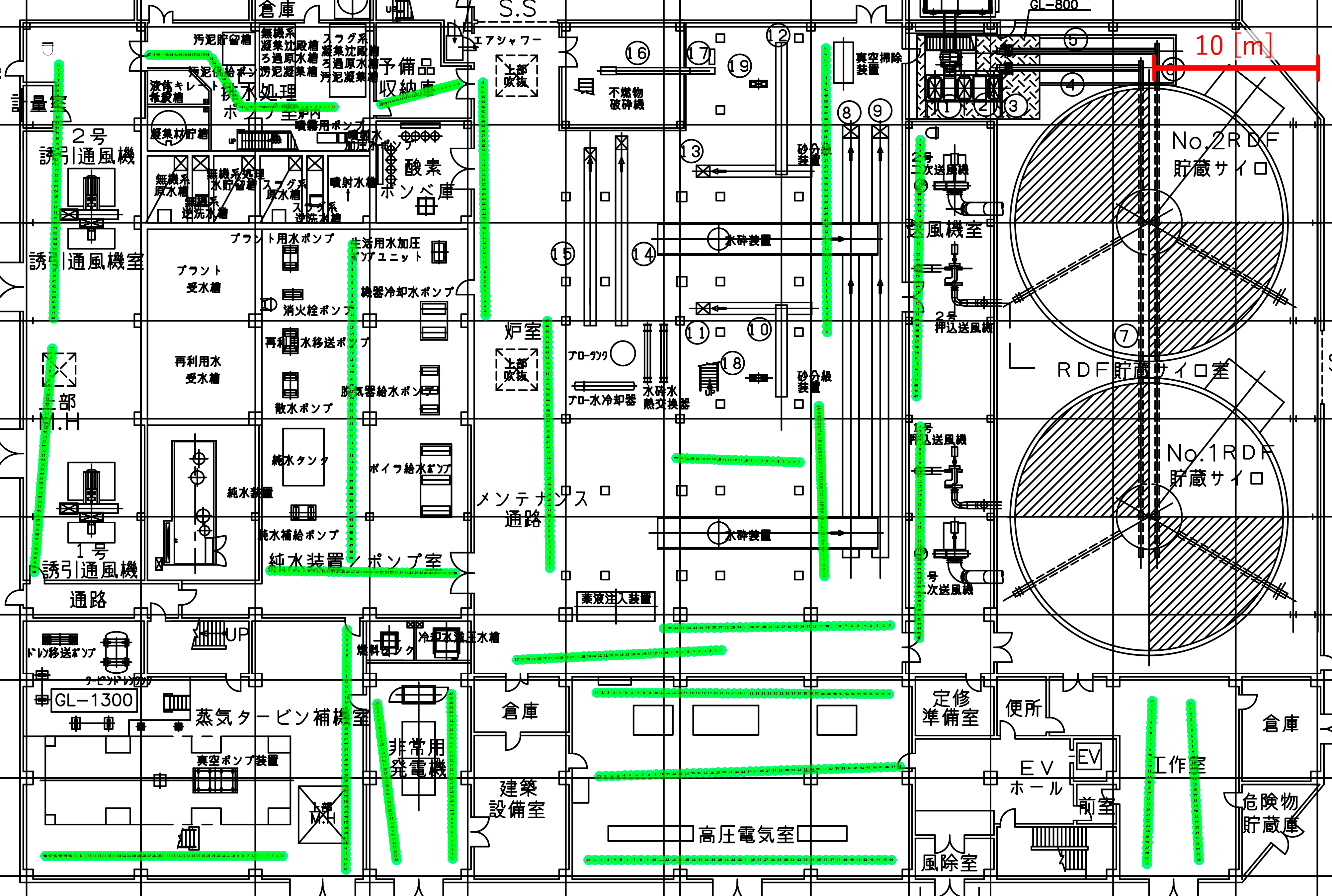}
    \caption{Collection points for environmental data during the offline phase.}
    \label{fig:train_pos}
\end{figure}

The research field is the Ishikawa Hokubu RDF Center, located in Ishikawa Prefecture, Japan.
This plant melts and processes Refuse Derived Fuel (RDF)
\footnote{
    {\bf R}efuse {\bf D}erived {\bf F}uel: RDF Refers to fuel produced from waste materials. Combustible materials such as paper, cloth, wood, and plastic that have not been recycled are crushed, dried, sorted, and formed into fuel.
}
produced at other processing plants. The environmental data used in this paper were acquired on September 13, 2021, at the first floor level of this incineration plant for the offline and estimation phases. The number of reference points is 1024 for the offline phase and 24 paths for the estimation phase. We collected the environmental data by placing a rope marked every 30 cm on each path and moving a cart along it. Therefore, the points in the paths are 30 cm apart from each other, and the total distance is 3072 cm. Figure~\ref{fig:train_pos} is a drawing of the research field showing the reference points and paths acquired for the offline phase. The number of points in the paths is 20 to 50 for the offline phase and estimation phase.

In Point Matching, for each 1024 ${\rm tar}(Mv, \ Mh)$ point and 1024 ${\rm ref}(Mv, \ Mh)$ point are compared.
In Path Matching and DTW Matching, the length $M$ of the path is the same. $TarPath[ \ ]$, is fixed to 20,
the minimum number of points in the path. As a result, $RefPoints[1:20]$ and $TarPath[1:20]$ are generated in 568 different ways.
In addition, $RefPoints[1:20]$ and $TarPath[1:20]$ are generated from both ends of the path.
That means considering the round trip of the path, resulting in a dataset of 1136 datasets each.

\section{Experimental Results and Discussions}
This chapter describes the results of applying the method described in Section \ref{sec:suggestion} to the data shown in Section \ref{sec:experimental_setup}.
We used an Apple Mac Pro (Late 2013, Xeon~E5 6~Cores / 12~Threads 3.5~GHz CPU, 32~GB 1866~MHz DDR3 memory) for the calculations and Implemented these algorithms by Python Ver. 3.8.12 and Numpy Ver. 1.21.2.

First, we compare the computation time of each method. Table~\ref{tab:calculation-time} shows the time taken to compute each method. This table shows that while Point Matching and Path Matching can be done within 1 second, DTW Matching takes more than 30 seconds. The reason for this is that the computational complexity of the DTW algorithm is $O(NM)$, which is larger than that of the other methods.

\begin{figure}[htbp]
    \centering
    \includegraphics[width=\linewidth]{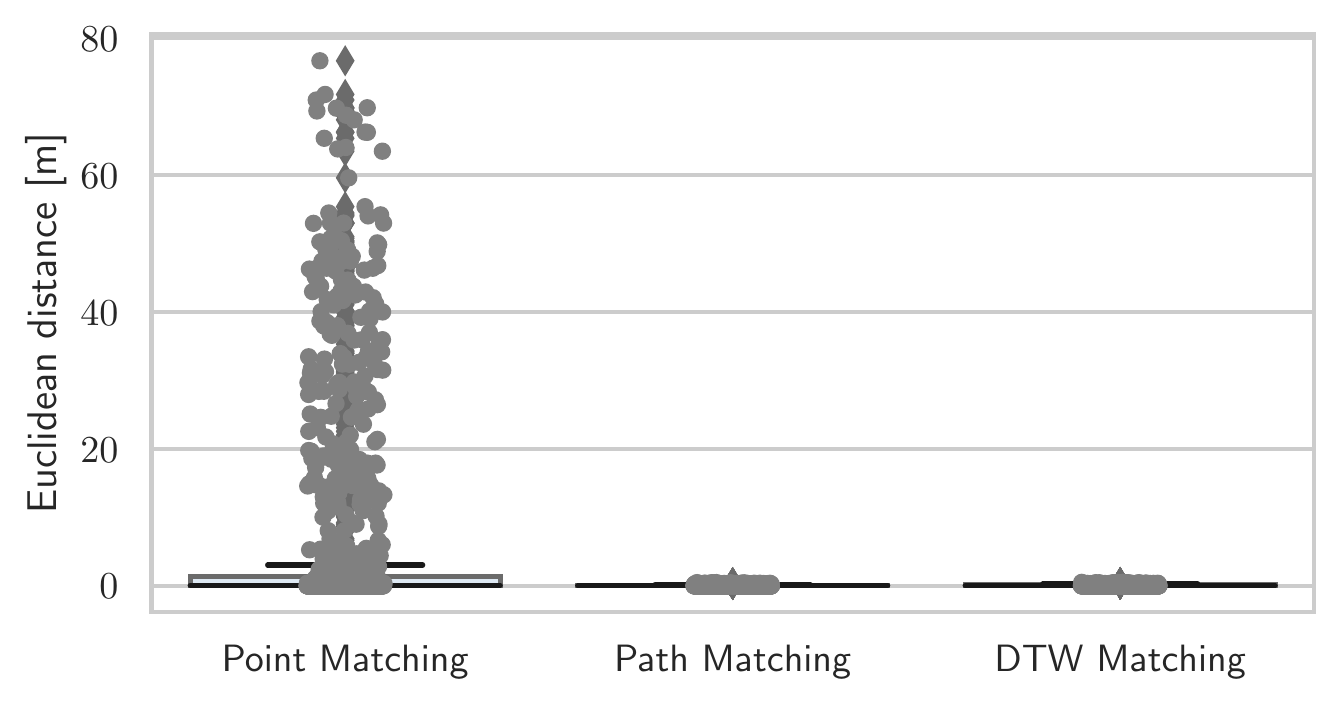}
    \caption{Comparison of the positioning error for each proposed method.}
    \label{fig:boxplot_123}
\end{figure}

Second, we compare the positioning error of each method by calculating the positioning error that differs among Point, Path, and DTW Matching. In Point Matching, the positioning error is defined as the Euclidean distance between the estimated target coordinates ${\rm tar}(x, y)$ and the estimated result coordinates ${\rm est}(x, y)$. In Path Matching and DTW Matching, the positioning error between the estimated target path $TarPath[1:20]$ and the estimated resultant path $EstPath[1:20]$ is calculated from the following equation~\eqref{eq:error}.
\begin{flalign}
    &TarPath[1:20] = (Tx_1, \ Ty_1), \ \dots, \ (Tx_{20}, \ Ty_{20}) \notag &\\
    &EstPath[1:20] = (Ex_1, \ Ey_1), \ \dots, \ (Ex_{20}, \ Ey_{20}) \notag &\\
    &error = \frac{1}{20}  \sum^{20}_{i = 1} \sqrt{(Tx_i - Ex_i)^2 + (Ty_i - Ey_i)^2} \label{eq:error} &
\end{flalign}
Figure~\ref{fig:boxplot_123} shows the positioning error of each proposed method calculated by the above method. And Table~\ref{tab:result-error} shows the values at the border of each interval.
From these results, we can see that Path Matching and DTW Matching are more accurate than Point Matching. This result can be assumed that a path can recognize the magnetic distribution more correctly in the matching process.

\begin{table}[htbp]
    \centering
    \caption{Calculation time}
    \label{tab:calculation-time}
    \begin{tabular}{rrr}
    \toprule
    Point Matching [s] & Path Matching [s] & DTW Matching [s] \\
    \midrule
    0.548 & 0.788 & 36.760 \\
    \bottomrule
    \end{tabular}
\end{table}

\begin{table}[htbp]
    \centering
    \caption{Estimation Error}
    \label{tab:result-error}
    \begin{tabular}{crrr}
    \toprule
    \multirow{2}*{Border} & Point Matching & Path Matching & DTW Matching \\
    & [m] & [m] & [m] \\
    \midrule \midrule
    100\% (max) & 76.70 & \textbf{0.46} & 0.50 \\
    75\% & 1.34 & 0.08 & 0.10 \\
    50\% (mid) & 0.00 & 0.00 & 0.00 \\
    25\% & 0.00 & 0.00 & 0.00 \\
    0\% (min) & 0.00 & 0.00 & 0.00 \\ \midrule
    mean & 6.89 & \textbf{0.05} & 0.06 \\ \bottomrule
    \end{tabular}
\end{table}

\begin{figure}[htbp]
    \centering
    \includegraphics[width=\linewidth]{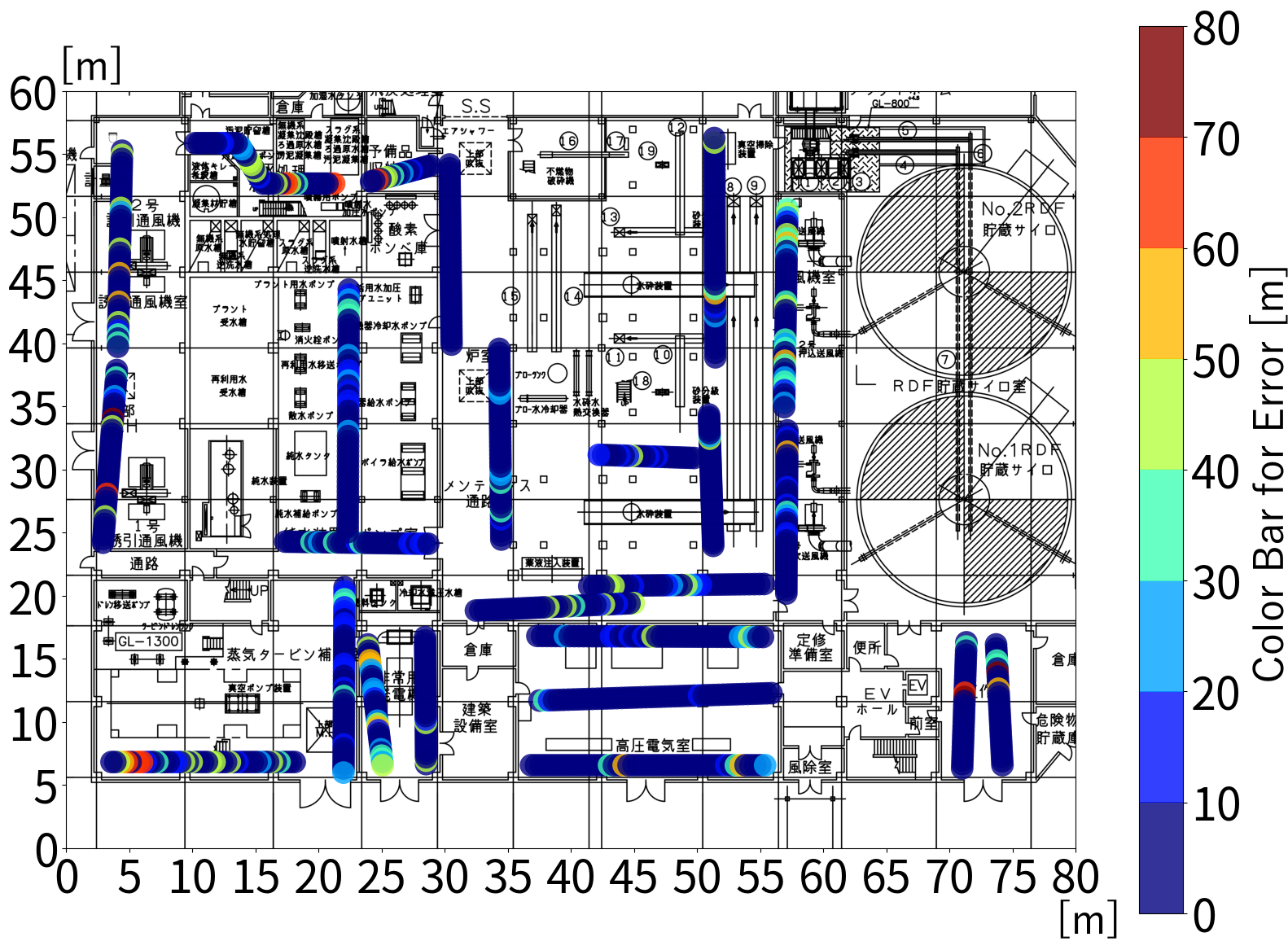}
    \caption{Heatmap of estimation results of Point Matching}
    \label{fig:heatmap_method1}
\end{figure}

Figure~\ref{fig:heatmap_method1} shows a heatmap of the positioning errors obtained at each reference point. This figure shows the positioning error in color according to the color bar on the right.
While most of the reference points are blue, but orange and red are widely spread, indicating that reference points with similar $M_v$ and $M_h$, which are compared in point matching, are widely dispersed. In Path Matching and DTW Matching, the maximum value is less than 1 m from Table~\ref{tab:result-error}, indicating that the positioning accuracy is significantly improved compared to Point Matching. This can be attributed to the fact that multiple physically close points are matched as a path, which reduces the influence of positioning errors at specific points observed in the first method.
On the other hand, we should note that the fingerprint spacing is 30 cm in each offline/online phase, which is an overly favorable condition compared to the actual one.

\begin{figure}[htbp]
    \centering
    \includegraphics[width=\linewidth]{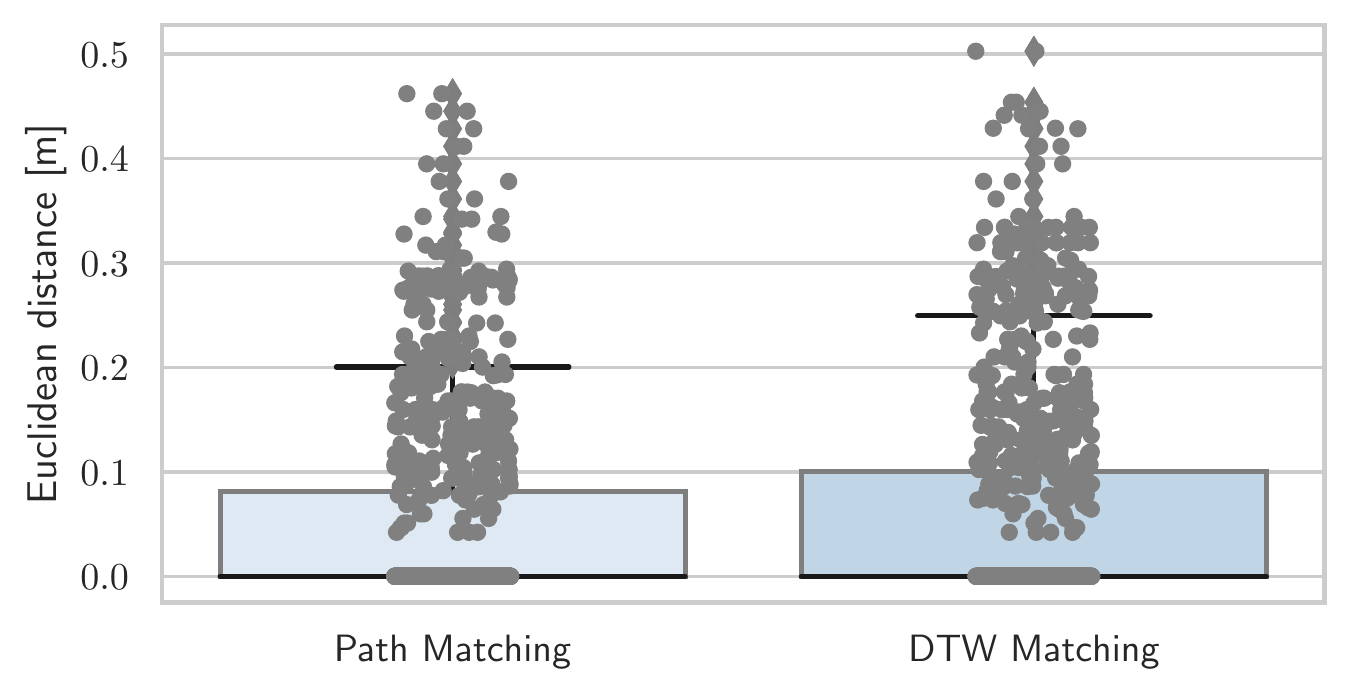}
    \caption{Comparison of positioning error between Path Matching and DTW Matching.}
    \label{fig:boxplot_23}
\end{figure}

\begin{figure}[htbp]
    \centering
    \includegraphics[width=\linewidth]{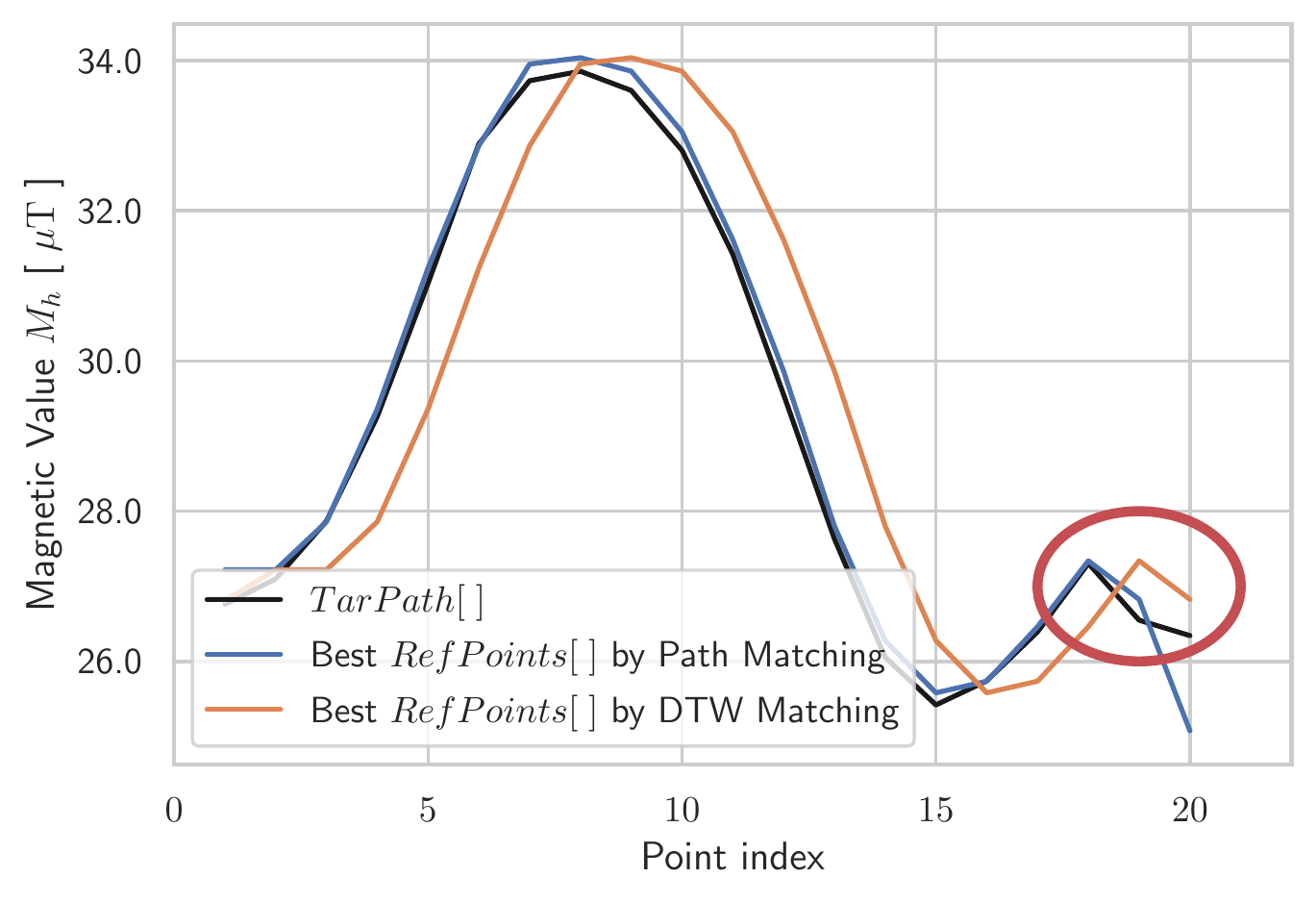}
    \caption{Estimated $M_v$ by Path Matching and DTW Matching.}
    \label{fig:magv}
\end{figure}

Furthermore, in Figure~\ref{fig:boxplot_23}, which compares the positioning errors of only Path Matching and DTW Matching, the box and whiskers of Path Matching are slightly lower than DTW Matching.
That is thought to be due to time warping in DTW.
Figure~\ref{fig:magv} shows an example of $M_v$ in $TarPath$, estimated by Path Matching and DTW Matching.
DTW is tolerant to changes in time, which is one reason why it matched $RefPoints[\;]$ with similar shapes, as shown in the red oval in Figure~\ref{fig:magv}.

\section{Conclusion}
Although workers can operate the current incineration plants remotely, workers still need to visit the site to inspect and clean the equipment. Moreover, when the number of workers is reduced to the minimum to reduce operation costs, only one worker must frequently visit the site. Therefore, a system to monitor the location of workers in real-time is required to detect unexpected accidents quickly. In this study, we are developing a system that uses magnetic FP for positioning. In this paper, we report on the current status of data collection and its analysis, which are necessary to implement the FP method using magnetic sensing in a vast area such as an incineration plant.

We also proposed three methods for FP matching: Point Matching, Path Matching, and DTW Matching. In these methods, the average positioning errors were 6.89 m, 0.05 m, and 0.06 m, respectively, when collecting the environmental data only with a measuring cart.

In the future, we will evaluate the accuracy of using the environmental data obtained without using a measuring cart in the estimation phase and examine the area division process, which is important for positioning across multiple floors, for verification closer to the actual operation.

\section*{Acknowledgment}
We would like to thank the staff of Ishikawa Hokubu RDF Center for their cooperation in data collection.

\bibliographystyle{IEEEtran}
\bibliography{mybib}

\end{document}